%% file: templateArxiv.tex
\documentclass{article}

\usepackage{PRIMEarxiv}

\usepackage[utf8]{inputenc} 
\usepackage[T1]{fontenc}    
\usepackage{hyperref}       
\usepackage{url}            
\usepackage{booktabs}       
\usepackage{multirow}
\usepackage{amsfonts}       
\usepackage{nicefrac}       
\usepackage{microtype}      
\usepackage{lipsum}
\usepackage{fancyhdr}       
\usepackage{graphicx}       
\usepackage{amsmath}
\graphicspath{{media/}}     

\usepackage{algorithm}%
\usepackage[noend]{algpseudocode}%

\title{Visualizing Latent Phase Structures in Locomotion Policies: A Multi-Environment Study with Temporal Feature Extension
}

\author{
  Daisuke Yasui*, Toshitaka Matuki, Hiroshi Sato \\
  Mathematics and Computer Science \\
  National Defense Academy of Japan \\
  Yokosuka, Japan\\
  \texttt{*ed24003@nda.ac.jp} \\
}

\begin{document}
\maketitle

\begin{abstract}
Deep reinforcement learning (DRL) has been shown to achieve high performance on locomotion control tasks in MuJoCo benchmarks such as HalfCheetah, Ant, and Walker2D.
However, visualizing the motion structures internally obtained by a trained policy function implemented as a deep neural network remains challenging.
It is known from biomechanics and related fields that locomotion control is realized through the repetition of motion phases such as the stance phase and swing phase.
In this study, we propose a framework for uncovering latent motion phase structures from trajectories generated by locomotion control policies through interaction with the environment.
The proposed method extends the clustering features from state observations alone to augmented features including actions, next states, and next actions, and introduces a method for determining the number of clusters that suppresses self-transitions.
Applying the proposed method to three environments---Ant-v5, HalfCheetah-v5, and Walker2D-v5---we successfully identified phase structures with clearer and more regular transition rules than those obtained by the existing method.
\end{abstract}


\input{01_introduction}
\input{02_relatedworks}
\input{03_evaluation}
\input{04_proposed_method}
\input{05_experiment_and_disucussion}
\input{06_conclusion}

\section*{Data Availability}
    The datasets and code used during the current study are available from the corresponding author on reasonable request.

\section*{Funding}
    This work was supported by JSPS KAKENHI Grant Numbers JP22K17969.
\bibliographystyle{plain}
\bibliography{sample}

\end{document}

%% file: 01_introduction.tex
\section{Introduction}
Deep reinforcement learning (DRL) has been shown to achieve high performance on locomotion control tasks, including MuJoCo benchmarks~\cite{Todorov2012MuJoCo} such as HalfCheetah, Ant, and Walker2D~\cite{Lillicrap2015Continuous}.
However, visualizing the motion structures internally obtained by a trained policy function implemented as a deep neural network is not straightforward.
It is known from biomechanics and robotics that locomotion control is realized through the repetition of motion phases such as the stance phase and swing phase~\cite{Collins2005EfficientBiped,Hurmuzlu2004HybridDynamics}.
This suggests that locomotion control policies trained via DRL may also autonomously acquire phase structures internally.

Yasui et al. verified this hypothesis in the HalfCheetah environment and proposed a method for identifying locomotion phase structures by embedding state sequences obtained through environment interaction into a low-dimensional space and applying clustering~\cite{yasui_XAI2026}.
However, directly applying the same method to different environments such as Ant-v5 and Walker2D-v5 fails to yield consistent phase structures.
This is attributed to two limitations of the prior method.
First, clustering based solely on state features, as in prior work, cannot fully capture the temporal role of each step.
This may cause two steps with similar postures but different subsequent transitions to be incorrectly merged into the same phase.
Second, there is no penalty for self-transitions, where the destination phase is the same as the current phase.
This may lead to merging distinct motion phases into the same phase because $H_c$ can be trivially minimized by inflating self-transitions, which makes the transition destination artificially predictable.

To address these issues, this study identifies phase structures across a broader range of locomotion control tasks than prior work by introducing the following two measures.
The first is to perform clustering based on features that concatenate not only the state of each step but also the action, next state, and next action, thereby appropriately distinguishing situations where state values are similar but the learned policy makes different decisions.
The second is to determine the number of clusters with a constraint on self-transitions, preventing distinct motion phases from being merged into the same phase.
The main contributions of this paper are as follows.
First, we propose a phase identification method that combines features obtained by concatenating actions, next states, and next actions with state features, along with clustering based on transition destination uniqueness and a penalty for self-transitions.
Second, we perform post-hoc visualization of phase structures in morphologically diverse MuJoCo locomotion environments that prior work could not handle, revealing environment-specific phase structures.

%% file: 02_relatedworks.tex
\section{Related Work}
\subsection{Reinforcement Learning and Policy Interpretation in Continuous 
Control}
Reinforcement learning (RL) is a framework in which an agent learns a policy that maximizes returns through interaction with the environment.
For continuous control tasks involving continuous state and action spaces, deep reinforcement learning (DRL) using deep neural networks is widely used~\cite{SuttonBarto2018,VanHasselt2016DoubleQ}.
As representative methods, Deterministic Policy Gradient (DPG) and its deep extension Deep Deterministic Policy Gradient (DDPG), have been proposed and demonstrated high performance in continuous action spaces~\cite{Lillicrap2015Continuous, lillicrap2019continuouscontroldeepreinforcement}.
Subsequently, Twin Delayed Deep Deterministic Policy Gradient (TD3) was proposed to mitigate the learning instability and overestimation bias of DDPG~\cite{Fujimoto2018TD3}.
TD3 demonstrates excellent performance on continuous control benchmarks represented by MuJoCo and has been widely adopted as one of the standard methods for robot locomotion and running control tasks.
In these methods, the policy function is represented as a multi-layer neural network that takes a state as input and directly outputs a continuous-valued action.
However, its internal decision-making process is a black box, making it difficult for humans to grasp the overall picture of the acquired representations.

To address this, research efforts have been made to visualize and interpret the overall structure of behaviors acquired by DRL agents.
Zahavy et al. embedded the internal state representations of an agent trained on Atari games into a low-dimensional space and visualized action modes and strategic state groups through clustering~\cite{Zahavy2016Graying}.
This approach is significant in demonstrating that semantically distinct behavioral states can exist within a policy function.
However, the resulting clusters did not explicitly consider temporal transition structures, and it was not possible to visualize the ordering relationships or periodicity among states.

\subsection{Visualization Within Policy Functions for Locomotion Control Tasks}
\begin{figure}[ht]
    \centering
    \includegraphics[width=0.5\columnwidth,page=7, trim={115mm 60mm 105mm 28mm}, clip]{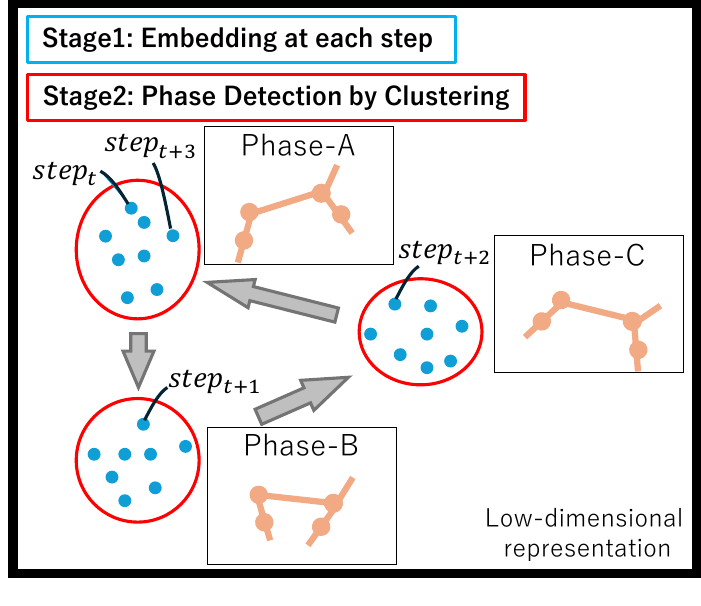}
    \caption{Overview of the proposed analysis framework. Stage 1 embeds each 
    step into a low-dimensional space based on motion features. Stage 2 
    identifies phases by clustering the embedded steps.}
    \label{fig:analysis_method}
\end{figure}

In locomotion control, it is known from biomechanics that motion is realized through the repetition of motion phases such as the stance phase and swing phase~\cite{Collins2005EfficientBiped,Hurmuzlu2004HybridDynamics}.
Yasui et al. hypothesized that a reinforcement learning policy function internally acquires representations that implicitly distinguish multiple motion phases, and outputs corresponding actions for each~\cite{yasui_XAI2026}.
They then proposed a method for visualizing the locomotion phase structures acquired by a policy function trained on the HalfCheetah task. 
An overview of this phase identification method is shown in Fig.~\ref{fig:analysis_method}.
The phase structure referred to here means that when multiple steps with similar motion features such as joint angles and torques are clustered into a single phase, the phases exhibit a consistent cyclic transition structure as illustrated in Fig.~\ref{fig:analysis_method}.

To identify the phase structure, as shown in Stage 1 of Fig.~\ref{fig:analysis_method}, the state $\mathbf{s}_t$ 
from the motion features
of each step was embedded into a 2-dimensional representation $z_t \in \mathbb{R}^2$ using UMAP~\cite{McInnes2018UMAPJOSS}, so as to preserve neighborhood relationships in the high-dimensional space.
The phase structure was then identified using hierarchical clustering that searches for the number of clusters at which the phase structure is most apparent, using the conditional entropy $H_c$:
\begin{equation}
  H_c = -\sum_{i=1}^{K} \frac{N_i}{N} \sum_{j=1}^{K} \frac{N_{ij}}{N_i} 
  \log \frac{N_{ij}}{N_i}
\end{equation}
where $K$ is the number of clusters, $N = \sum_{k=1}^{K} N_k$ is the total number of steps, $N_i$ is the number of steps belonging to cluster $i$, and $N_{ij}$ is the number of transitions from cluster $i$ to cluster $j$.
A smaller $H_c$ means that the next cluster is more predictable from the current cluster.
The optimal number of clusters $K^*$ is determined by $K^* = \mathop{\arg\min}_{K \in \{2,\ldots,20\}} H_c(K)$.
Note, however, that $H_c$ can also be minimized when there are many self-transitions to the same phase.

However, this work is limited to HalfCheetah as the target of visualization, and has not succeeded in identifying phase structures with the properties described in the next section for locomotion environments such as Ant and Walker2D.

%% file: 03_evaluation.tex
\section{Evaluation Methodology for Phase Visualization and Limitations of the Existing Method}\label{sec:evaluation}

The validity of the identified phase structure can be evaluated from two perspectives.
The first perspective is geometric properties.
If the phases have a consistent cyclic structure, the trajectories representing phase transitions should also be preserved in the embedding space.
Furthermore, if adjacent phases with similar motion features are also placed in proximity in the embedding space, the amount of change in motion features between adjacent phases becomes roughly uniform.
Therefore, if the number of phases is $K$, the phase transitions are expected to appear as a trajectory approximating a regular $K$-gon with a consistent direction of rotation.
Taking this as the ideal state, we can evaluate how much the actual embedding deviates from this shape.
However, existing methods may not sufficiently achieve this property across a wide range of locomotion tasks.
One possible explanation is that existing methods perform clustering based solely on the similarity of state features; when small differences in state lead to large changes in action, steps in a single phase may transition to multiple different phases.
To address this, it is necessary to perform clustering by including next states and next actions as features as well.

The second perspective on phase structure validity is transition regularity.
If the transition destination from steps belonging to the same phase is uniquely determined, it means that transitions occur regularly, and this indicates that a consistent cyclic transition structure has been captured.
In the existing method, clustering is performed to minimize $H_c$; however, $H_c$ can be minimized even when there are many self-transitions to the same phase.
As a result, coarser clustering is promoted, and there is a possibility that a phase structure that sufficiently captures regular transitions may not be identified.
To address this, it is necessary to perform clustering that simultaneously evaluates both the uniqueness of the transition destination and the infrequency of self-transitions.

%% file: 04_proposed_method.tex
\section{Proposed Method}
We propose a method that addresses the issues raised in the previous section as follows.
The existing method used only the state feature $\mathbf{s}_t$ at each time step $t$ as the clustering feature.
In this study, we extend this by using the feature composition $[\mathbf{s}_t,\; \mathbf{a}_t,\; \mathbf{s}_{t+1},\; \mathbf{a}_{t+1}]$, which additionally includes the action, next state, and next action.
Each element is independently z-score normalized, concatenated, and scaled by $1/\sqrt(d)$ (where $d$ is the dimensionality) to prevent variance inflation as the dimensionality increases. 
Based on these features, embedding into a 2D space is performed using UMAP~\cite{McInnes2018UMAPJOSS}.

In the resulting embedding space, clustering is attempted for cluster counts $K$ ranging from $2$ to $20$ using hierarchical clustering with Ward's method, and the cluster count $K^*$ that maximizes the objective function is found using the elbow method.
We define the external concentration $\mathrm{C_{ext}}$ as the objective function for the search.
$\mathrm{C_{ext}}$ is defined using the transition probability $P_{ij}$ from cluster $i$ to $j$ and the self-transition probability $P_{ii}$ as follows:
\begin{equation}
  \mathrm{C_{ext}}(K)
  = \sum_{i=0}^{K-1} w_i \cdot (1 - P_{ii})
    \cdot \max_{j \neq i} {P_{ij}}
    \label{eq:Cext}
\end{equation}
where $w_i$ is the relative frequency of cluster $i$.
$(1 - P_{ii})$ is the proportion of external transitions, and $\max_{j \neq i} P_{ij}$ is the concentration of external transition destinations.
A larger value of this product means that external transitions are more frequent and the destination is more uniquely determined, indicating a clearer phase structure.
To select the number of clusters $K$, the elbow method on $\mathrm{C_{ext}}$ is used to suppress unnecessary increases in the number of clusters:
\begin{equation}
  K^* = \operatorname*{argmax}_{K \in \{2, 3, ..., 20\}}
        \left(\widetilde{\mathrm{C_{ext}}}(K) - \tilde{K}\right)
\end{equation}
where $\tilde{\cdot}$ denotes MinMax normalization of each variable to the range $[0,1]$.
Specifically, the normalization is performed as:
\begin{equation}
    \tilde{C}_{\text{ext}}(K) = \frac{C_{\text{ext}}(K) - \min_K 
    C_{\text{ext}}}{\max_K C_{\text{ext}} - \min_K C_{\text{ext}}}
\end{equation}
\begin{equation}
    \tilde{K} = \frac{K - K_{\min}}{K_{\max} - K_{\min}}
\end{equation}

%% file: 05_experiment_and_disucussion.tex
\section{Experiment and Discussion}
Three morphologically distinct environments from the MuJoCo locomotion control benchmark~\cite{Todorov2012MuJoCo}---Ant-v5, HalfCheetah-v5, and Walker2D-v5---were used in the experiments.
For each environment, a policy pre-trained with the TD3 algorithm was used, and state and action sequences from 5 episodes of 1000 steps each were  collected to compare the results of the proposed and existing methods.
In this section, we first perform a qualitative comparison of the phase structure visualization results in each environment, followed by a quantitative comparison.
The hyperparameters for model training, UMAP embedding, and clustering follow those of 
Yasui et al.\cite{yasui_XAI2026}.
We then analyze the phase transition matrices to identify the dominant phase transitions and confirm the robot's motion during those transitions.
Finally, we conduct an ablation study to analyze the effects of the two measures introduced by the proposed method.

\subsection{Visualization Experiment}
    \begin{figure}[t]
        \centering
        \includegraphics[width=0.8\columnwidth,page=12, trim={50mm 50mm 135mm 15mm}, clip]{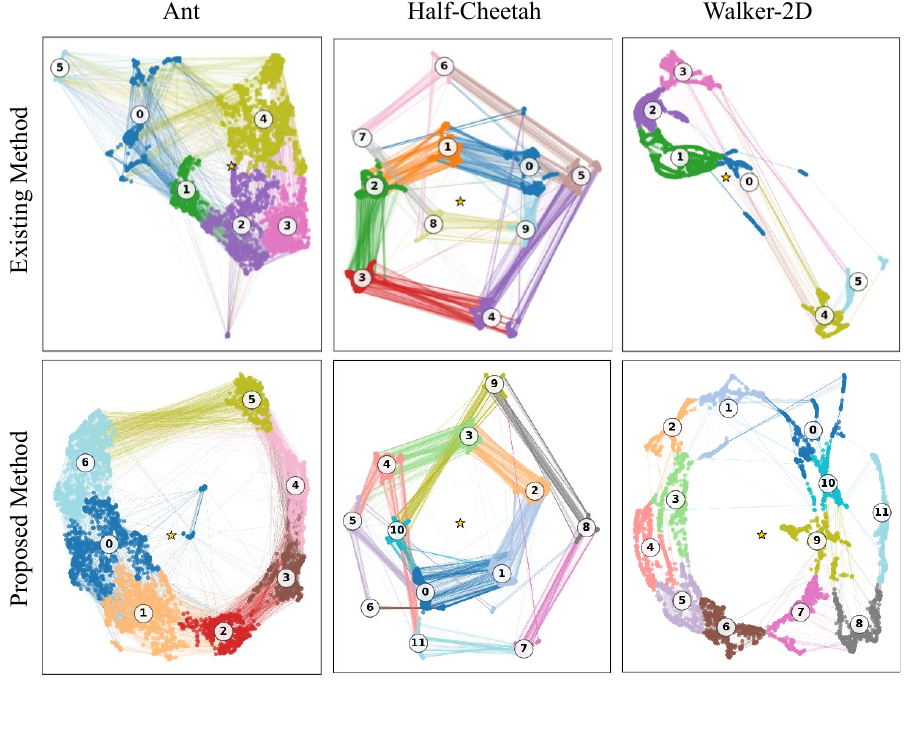}
        \caption{Visualization results of the proposed and existing methods across three environments. Each node represents a step embedded in 2D space, colored by cluster assignment. Edges represent transitions between steps, colored by the source cluster. The star marker represents the centroid of all steps.}
        \label{fig:visualize}
    \end{figure}

    Fig.~\ref{fig:visualize} shows the results of embedding each step into a 2-dimensional space and applying clustering using the existing and proposed visualization methods, based on the features of each step obtained through interaction with the environment in each locomotion control task.
    Each node represents a step in the 2-dimensional space, and the star marker represents the centroid of all steps.
    Edges represent transitions between steps; the color of each node indicates the phase it belongs to, and the color of each edge indicates the source phase.
    Numbers indicate phase labels, and the position of each number corresponds to the centroid of the cluster associated with that phase.

    Looking at the overall visualization results, when transitioning from one phase to another, transitions consistently occur toward specific phases, suggesting a cyclic pattern.
    This indicates that phase transitions follow a regular cyclic order rather than occurring randomly.
    Furthermore, it can be visually observed that in the proposed method, inter-phase transitions are arranged in a more annular pattern centered on the star marker representing the overall centroid.
    This annular arrangement of phase transitions geometrically reflects the periodic phase transition structure of locomotion, allowing one to intuitively confirm the structure in which each phase cycles in a consistent order.
    On the other hand, it is difficult to visually judge which method is superior in terms of transition regularity, and quantitative evaluation is necessary.

\subsection{Quantitative Evaluation of Visualization Results}
    We now perform a quantitative evaluation of the visualization results presented in the previous section.
    The silhouette score~\cite{rousseeuw1987silhouettes} is a common metric for evaluating distance-based clustering performance.
    The silhouette score is defined as follows, with N denoting the total number of samples (steps): 
    \begin{equation}
        \text{Silhouette} = \frac{1}{N} \sum_{i=1}^{N} \frac{b(i) - a(i)}{\max(a(i),\ b(i))}
        \label{shihouette}
    \end{equation}
    where $a(i)$ is the mean distance between sample $i$ and all other samples in the same cluster, $b(i)$ is the mean distance between sample $i$ and the samples in the nearest neighboring cluster, and a larger value indicates that clusters are more mutually separated.
    The range is $[-1, 1]$.

    As can be seen from Eq.~\ref{shihouette}, the silhouette score is based solely on distances between samples and does not directly consider transition relationships.
    The clustering targets in this study are steps that vary continuously along trajectories, and temporally close steps are also placed close together in the embedding space.
    Consequently, increasing the number of clusters $\bar{K}$ partitions the trajectory more finely, and the distance $b(i)$ to the nearest neighboring cluster tends to shrink.
    As a result, independently of whether clusters are valid as motion phases, coarser partitions with smaller $\bar{K}$ tend to yield higher silhouette scores.
    Table~\ref{table:result} shows that, compared to the existing method, the proposed method yields lower silhouette scores for Ant and Walker2D (where $\bar{K}$ is substantially larger), and comparable scores for HalfCheetah (where $\bar{K}$ is similar). 
    These results are consistent with the aforementioned bias with respect to $\bar{K}$.
    Thus, evaluation metrics that do not directly consider transition relationships may be strongly biased by the number of clusters itself.
    More fundamentally, what matters in motion phase identification, as discussed in Sec.~\ref{sec:evaluation}, is whether each phase forms a geometrically consistent cyclic structure and exhibits regular transitions.
    Based on these considerations, we introduce metrics that quantify geometric properties and transition regularity based on inter-step transition relationships.

    We quantitatively analyze whether the proposed method achieves a more annular arrangement of phase transitions compared to the existing method, and whether it exhibits higher uniqueness of transition destinations with fewer self-transitions.

    First, we evaluate whether the phase transitions form an annular arrangement.
    We define rotational regularity $\mathcal{R}$ as a metric to quantify the annular arrangement of phase transitions in the 2-dimensional embedding space.
    Let $\bar{e}$ be the centroid of all points in the embedding space, and $e_t$ be the embedding point at time $t$.
    When an inter-phase transition occurs between the step at time $t$ and the step at time $t+1$, the sign of the cross product of ``the current position as seen from the centroid'' $\boldsymbol{r}_t = e_t - \bar{e}$ and ``the transition direction'' $\boldsymbol{v}_t = e_{t+1} - e_t$ is positive if the transition is counterclockwise and negative if clockwise.
    $\mathcal{R}$ is defined as the absolute value of the mean of this cross product normalized by the vector magnitudes, over all between-cluster transitions:
    \begin{equation}
        \mathcal{R} = \left| \frac{1}{|\mathcal{T}|}
        \sum_{t \in \mathcal{T}}
        \frac{\boldsymbol{r}_t \times \boldsymbol{v}_t}
             {\|\boldsymbol{r}_t\| \cdot \|\boldsymbol{v}_t\|} \right|
        \label{eq:R}
    \end{equation}
    where $\mathcal{T}$ is the set of all transitions excluding self-transitions.
    $\mathcal{R}$ takes values in $[0, 1]$; a larger value indicates a stronger tendency for transitions to rotate in one direction, showing that clusters are arranged in an annular pattern.
    From Table~\ref{table:result}, the proposed method outperforms the existing method in $\mathcal{R}$ across all environments, quantitatively confirming that the transitions are arranged in a more annular pattern.

    Next, transition regularity is evaluated using the external concentration $C_{ext}$ from Eq.~\ref{eq:Cext}.
    From Table~\ref{table:result}, the proposed method outperforms the existing method in $C_\text{ext}$ across all environments, quantitatively confirming that transition regularity is higher than that of the existing method.

    \begin{table}[H]
    \centering
    \caption{Quantitative comparison of existing and proposed methods 
    (10 seeds mean).}
    \label{table:result}
    \resizebox{0.8\columnwidth}{!}{%
    \begin{tabular}{llccc}
        \hline
        & & \textbf{Ant} & \textbf{HalfCheetah} & \textbf{Walker2D} \\
        \hline
        \multirow{4}{*}{Existing}
          & $\bar{K}$ & $6.0 \pm 0.0$ & $8.0 \pm 1.8$ & $6.3 \pm 0.9$ \\
          & Silhouette ($\uparrow$) & $\mathbf{0.474 \pm 0.033}$ & $0.665 \pm 0.046$ & $\mathbf{0.543 \pm 0.060}$ \\
          & $\mathcal{R}$ ($\uparrow$) & $0.371 \pm 0.146$ & $0.614 \pm 0.206$ & $0.403 \pm 0.256$ \\
          & $\mathcal{C}_{\mathrm{ext}}$ ($\uparrow$) & $0.174 \pm 0.031$ & $0.775 \pm 0.099$ & $0.007 \pm 0.007$ \\
        \hline
        \multirow{4}{*}{Proposed}
          & $\bar{K}$ & $10.3 \pm 1.9$ & $7.7 \pm 2.0$ & $12.3 \pm 5.3$ \\
          & Silhouette ($\uparrow$) & $0.422 \pm 0.025$ & $\mathbf{0.693 \pm 0.055}$ & $0.493 \pm 0.034$ \\
          & $\mathcal{R}$ ($\uparrow$) & $\mathbf{0.670 \pm 0.292}$ & $\mathbf{0.756 \pm 0.191}$ & $\mathbf{0.476 \pm 0.209}$ \\
          & $\mathcal{C}_{\mathrm{ext}}$ ($\uparrow$) & $\mathbf{0.462 \pm 0.062}$ & $\mathbf{0.846 \pm 0.054}$ & $\mathbf{0.023 \pm 0.025}$ \\
        \hline
        \end{tabular}
        }
    \end{table}

\subsection{Comparison of Transition Matrices}
    \begin{figure}[t]
        \centering
        \includegraphics[width=0.8\columnwidth,page=13, trim={50mm 50mm 135mm 15mm}, clip]{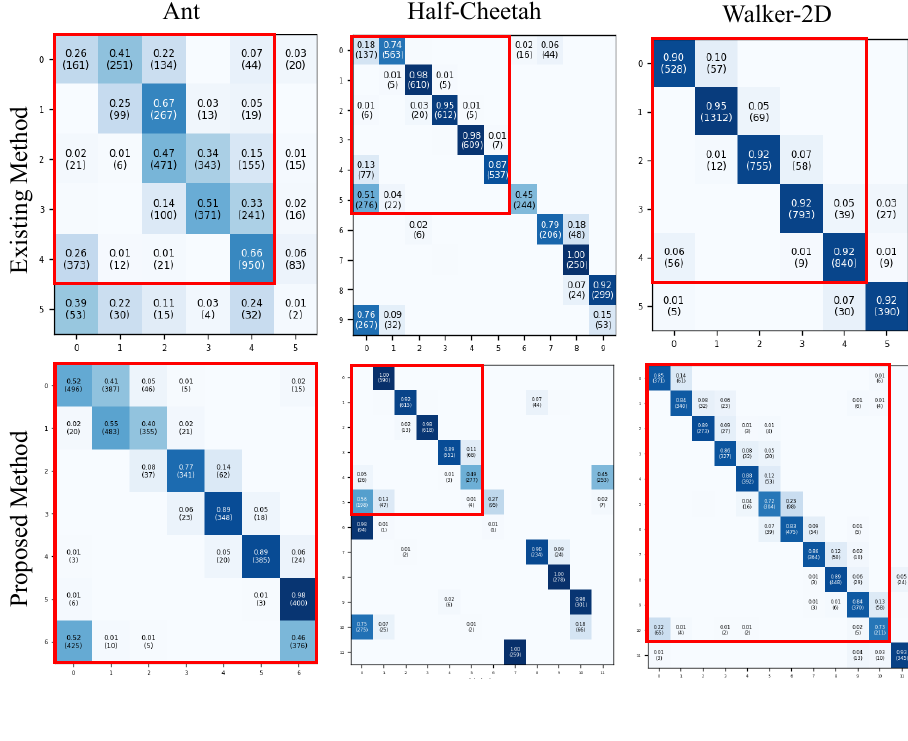}
        \caption{
        Transition probability matrices of the existing and proposed methods for each environment. 
        Rows and columns correspond to source and destination clusters, respectively. 
        Each cell shows the transition probability and (step count). The red border highlights the dominant transition path returning to Phase 0.
        }
        \label{fig:transition}
    \end{figure}

    In this section, we analyze the transition structure between clusters.
    Fig.~\ref{fig:transition} shows the transition matrices for the existing and proposed methods.
    Each row represents the source phase label and each column represents the destination phase label.
    The value in each cell represents the transition probability from the source phase to each destination phase, visualized using a heatmap.
    The number in parentheses indicates the number of steps that underwent that transition.

    Phase labels are assigned sequentially: for each labeled phase, the label of the next phase is given to the phase that receives its most frequent external transitions.
    If the phase with the highest external transition probability has already been labeled, the same labeling procedure is repeated starting from an unlabeled phase.
    The diagonal elements of the transition matrix represent self-transitions, and the cell adjacent to the diagonal represents the dominant external transition destination.
    The red border indicates the longest dominant transition pattern until the external transition destination returns to Phase 0.

    From Fig.~\ref{fig:transition}, it can be seen that the characteristics of the transition matrices differ across tasks.
    For example, in HalfCheetah, the diagonal elements representing self-transitions are extremely small compared to the other environments, and transitions are concentrated in the cells adjacent to the diagonal, indicating that the transition destination is roughly constant.
    This shows that each phase in HalfCheetah is represented by approximately a single step.
    In contrast, Ant and Walker2D have larger diagonal elements compared to HalfCheetah, with Walker2D having the largest proportion of diagonal elements.
    This indicates that the motion within each phase is represented by multiple steps.

    For Ant and HalfCheetah, the difference in the number of phases within the red-bordered dominant transition between the proposed and existing methods is approximately 1 to 2.
    For Walker2D, however, the proposed method yields more than twice as many phases in the dominant transition compared to the existing method.
    In Walker2D, the suppression of self-transitions by the proposed method revealed the complexity of external transition destinations, resulting in a finer-grained phase decomposition and a substantially larger number of phases than the existing method.

\subsection{Comparison of Rendering Results for Each Phase}
    \begin{figure}[ht]
        \centering
        \includegraphics[width=0.7\columnwidth,page=11, trim={95mm 20mm 105mm 30mm}, clip]{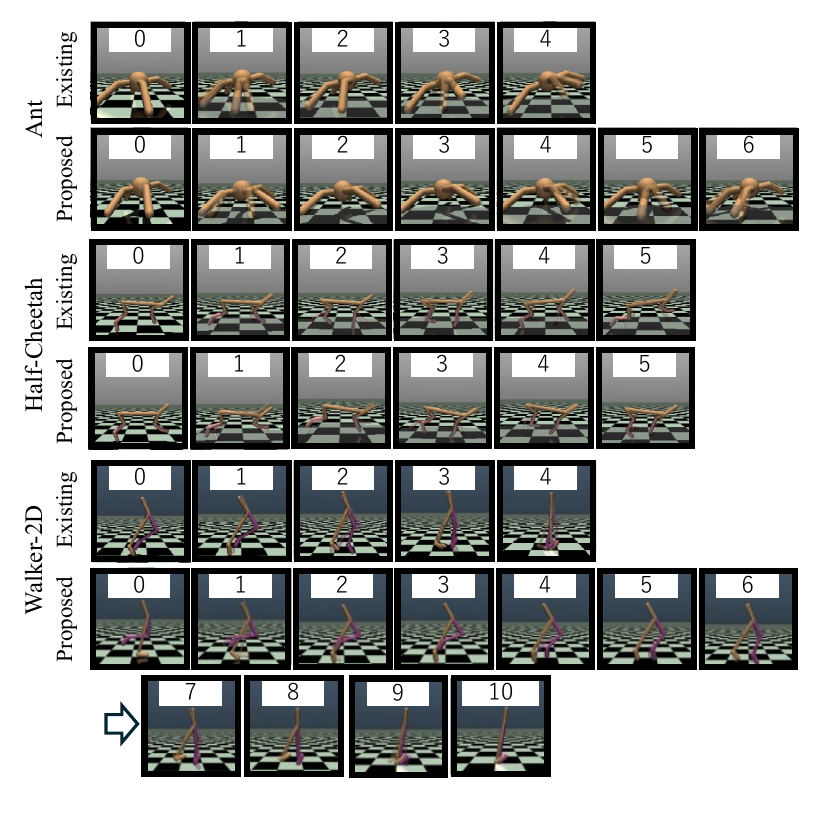}
        \caption{
        Rendering results of the dominant phase transitions for the existing and proposed methods in each environment. 
        Each image shows the robot posture at a given phase, with a transparent overlay of the previous step's posture to illustrate the motion between consecutive phases.
        }
        \label{fig:rendering}
    \end{figure}

    We visualize the robot postures during the dominant phase transitions highlighted by the red border in Fig.~\ref{fig:transition}.
    Fig.~\ref{fig:rendering} shows the rendering results of the robot state at each phase during the dominant transitions for both the existing and proposed methods across each environment.
    Each image shows the robot state at a given phase, with the transparent overlay representing the robot state in the previous step.
    This allows for an intuitive understanding of the specific motion from the previous phase that led to the current posture.

    In Ant, both the existing and proposed methods capture each phase as corresponding to leg extension, contraction, and weight shift, progressively reflecting the coordinated four-legged locomotion pattern.
    In HalfCheetah, it can be observed that landing, jumping, and aerial states are repeated.
    In Walker2D, for both the existing and proposed methods, the brown leg is dragged almost like a rod in all postures, while the purple leg repeats a cycle of pulling in, kicking out, and landing.
    The proposed method captures the leg cycle in finer detail through its more granular phase decomposition.

\subsection{Ablation Study}

\begin{table}[H]
    \centering
    \caption{Experimental results across environments and feature conditions(10 seeds mean).}
    \resizebox{0.7\columnwidth}{!}{%
    \begin{tabular}{cccccl}
    \hline
    \textbf{Env.} & \textbf{Feat.} & \textbf{Obj.} & \textbf{$\mathcal{R}$(↑)} & \textbf{$C_{ext}$(↑)} & \textbf{Cond.} \\
    \hline
    \multirow{4}{*}{Ant}
      & state & $H_c$        & $0.371 \pm 0.146$             & $0.174 \pm 0.031$             & Existing \\
      & state & $C_{ext}$    & $0.390 \pm 0.147$             & $0.244 \pm 0.039$             & + $C_{ext}$\\
      & all   & $H_c$        & $0.645 \pm 0.312$             & $0.298 \pm 0.061$             & + Feat.\\
      & all   & $C_{ext}$    & $\mathbf{0.670 \pm 0.292}$    & $\mathbf{0.462 \pm 0.062}$    & Proposed \\
    \hline
    \multirow{4}{*}{HalfCheetah}
      & state & $H_c$        & $0.614 \pm 0.206$             & $0.775 \pm 0.099$             & Existing \\
      & state & $C_{ext}$    & $0.613 \pm 0.208$             & $0.776 \pm 0.096$             & + $C_{ext}$\\
      & all   & $H_c$        & $\mathbf{0.762 \pm 0.172}$    & $0.834 \pm 0.068$             & + Feat.\\
      & all   & $C_{ext}$    & $0.756 \pm 0.191$             & $\mathbf{0.846 \pm 0.054}$    & Proposed \\
    \hline
    \multirow{4}{*}{Walker2D}
      & state & $H_c$        & $0.403 \pm 0.256$             & $0.007 \pm 0.007$             & Existing \\
      & state & $C_{ext}$    & $0.352 \pm 0.207$             & $0.018 \pm 0.015$             & + $C_{ext}$\\
      & all   & $H_c$        & $0.464 \pm 0.236$             & $0.008 \pm 0.008$             & + Feat.\\
      & all   & $C_{ext}$    & $\mathbf{0.476 \pm 0.209}$    & $\mathbf{0.023 \pm 0.025}$    & Proposed \\
    \hline
    \end{tabular}
    }
    \label{tab:ablation}
\end{table}

    We analyze the effects of the feature extension and the cluster count selection using $C_{ext}$ introduced in the proposed method.
    Table~\ref{tab:ablation} presents the results of comparing $\mathcal{R}$ and $C_{ext}$ across Ant, HalfCheetah, and Walker2D for the existing method and conditions where feature extension, the new objective function, or both, are added to the existing method.
    Compared to the method using only state features, the condition with extended features shows improvements in both $\mathcal{R}$ and $C_{ext}$.
    These results suggest that feature extension is effective for improving both $\mathcal{R}$ and $C_{ext}$.
    Compared to the method using $H_c$ as the clustering objective, the method using $C_{ext}$ tends to improve $C_{ext}$.
    On the other hand, the trend in $\mathcal{R}$ is unclear.
    Feature extension consistently improves $\mathcal{R}$ even when combined with $C_{ext}$, indicating that $C_{ext}$ does not cancel out the benefit of feature extension on $\mathcal{R}$.
    This suggests that while using $C_{ext}$ as the objective improves $C_{ext}$, it does not degrade $\mathcal{R}$ to the extent of negating the effect of feature extension.

%% file: 06_conclusion.tex
\section{Conclusion}
    In this study, we proposed a method for visualizing motion phase structures across morphologically diverse MuJoCo environments, based on trajectories generated through interaction between the environment and locomotion control policies trained via deep reinforcement learning.
    By extending the features and improving the objective function for cluster count selection relative to the existing method, we identified phase structures with fewer self-transitions arranged in a more annular pattern.
    Having succeeded in identifying phase structures even in morphologically diverse environments, we will focus our future work on interpreting the decision-making process within each phase and the overall phase structure.